\documentclass[letterpaper, 10 pt, conference]{ieeeconf}  

\IEEEoverridecommandlockouts                              

\usepackage{tabularx}
\usepackage{amsmath}
\usepackage{amssymb}
\usepackage{graphics}
\usepackage{graphicx}
\usepackage{multirow}

\usepackage{url}
\usepackage{xparse}
\usepackage{amssymb}
\usepackage{xspace}
\usepackage{nicefrac}
\usepackage{microtype}
\usepackage{amsthm}
\usepackage{amsfonts}
\usepackage{enumerate}
\usepackage{setspace}
\usepackage{xspace}
\usepackage{tabularx}
\let\labelindent\relax
\usepackage{enumitem}
\usepackage{wrapfig}
\usepackage{microtype}
\usepackage{graphicx}
\usepackage{subfigure}
\usepackage{booktabs} %
\usepackage{tikz}
\usepackage{hyperref}
\usepackage[style=ieee, backend=bibtex]{biblatex}
\usepackage[colorinlistoftodos]{todonotes}
\usepackage[style=ieee, backend=bibtex]{biblatex}
\newcommand{\aqt}[1]{\textcolor{blue}{#1}}

\newcommand{\soft}{\texttt{SOFT}$(\cdot)$\xspace}
\newcommand{\pvt}{\texttt{PVT}\xspace}
\newcommand{\softpvt}{\texttt{SOFT}$(\pvt)$\xspace}

\title{\LARGE \bf
Recasting Generic Pretrained Vision Transformers As Object-Centric Scene Encoders For Manipulation Policies
}
\author{Jianing Qian$^{1}$, Anastasios Panagopoulos$^{1}$ and Dinesh Jayaraman$^{1}$
\thanks{$^{1}$All authors are with GRASP Lab, Computer and Information Science Department,
        University of Pennsylvania, 3330 Walnut St, USA
        {\tt\small jianingq, anpans, dineshj@seas.upenn.edu}}%
}

\begin{document}

\maketitle
\thispagestyle{empty}
\pagestyle{empty}

\begin{abstract} 
Generic re-usable pre-trained image representation encoders 
have become a standard component of methods for many computer vision tasks. As visual representations for robots however, their utility has been limited, leading to a recent wave of efforts to pre-train robotics-specific image encoders that are better suited to robotic tasks than their generic counterparts.
We propose \emph{Scene Objects From Transformers}, abbreviated as \soft
, a wrapper around pre-trained vision transformer (\pvt) models that bridges this gap without any further training. Rather than construct representations out of only the final layer activations, \soft individuates and locates object-like entities from \pvt attentions, and describes them with \pvt activations, producing an object-centric embedding.  
Across standard choices of generic pre-trained vision transformers \pvt, we demonstrate in each case that policies trained on \softpvt far outstrip standard \pvt representations for manipulation tasks in simulated and real settings, approaching the state-of-the-art robotics-aware representations. Code, appendix and videos: https://sites.google.com/view/robot-soft/ 
\end{abstract}



\section{INTRODUCTION}
In the last decade, computer vision researchers have demonstrated consistently~\cite{sermanet2013overfeat, girshick2014rich} that visual features pre-trained with the right objectives on large image datasets offer large advantages for learning to perform new visual tasks.
Consequently, it is the standard practice today for generic ``foundation'' image encoder models~\cite{he2016deep, chen2020simple, chen2021exploring, grill2020bootstrap, chen2020improved, dosovitskiy2020image, caron2021emerging, Oquab2023DINOv2LR, he2021masked, Amir2021DeepVF} 
to be reused for virtually all standard computer vision tasks, including image classification, object detection, segmentation, pose estimation, monocular depth estimation, and video understanding, sometimes even despite domain differences with the pre-training data~\cite{kim2022broad}.
Thanks to this robust pre-training advantage, vision algorithms have been able to leverage the benefits of ever-improving neural network backbones, training algorithms, and pre-training objectives to boost performance across many tasks.

However, advantages from using these generic pre-trained image encoders have remained notably elusive in robotics
~\cite{zeng2018learning,yen2020learning}. In response, there has been a recent wave of image encoders pre-trained specifically for robotic tasks, especially manipulation, often exploiting the temporal structure of human videos of manipulation activities~\cite{nair2022r3m,ma2022vip,ma2023liv,radosavovic2023real,majumdar2023we}. These representations already achieve much more promising results than their generic counterparts for enabling robot learning and control, even though the pre-training datasets are less diverse, and objectives and architectures are less mature.  

Towards understanding and addressing the deficiencies of generic image encoders for robotics, we focus on the inference procedure that extracts image representations from such encoders to provide to downstream policies.
With no new training, we propose a procedure to extract a more robotics-relevant representation of an image from a simple forward pass through a generic pre-trained encoder.
Our insight is that the computational trace of a forward pass through a modern transformer neural network includes not just key, value, and query feature activations but also attention weights at each layer. Such attention weights have been noted in prior works to contain meaningful information, such as about referring expressions and phrases in language~\cite{vashishth2019attention,abnar2020quantifying}, and object regions in images~\cite{caron2021emerging}. However, to our knowledge, all prior attempts to apply pre-trained image encoders to downstream tasks have ignored this information, instead relying solely on activations.

\begin{figure*}[t]
\centering
\includegraphics[width=0.7\textwidth]{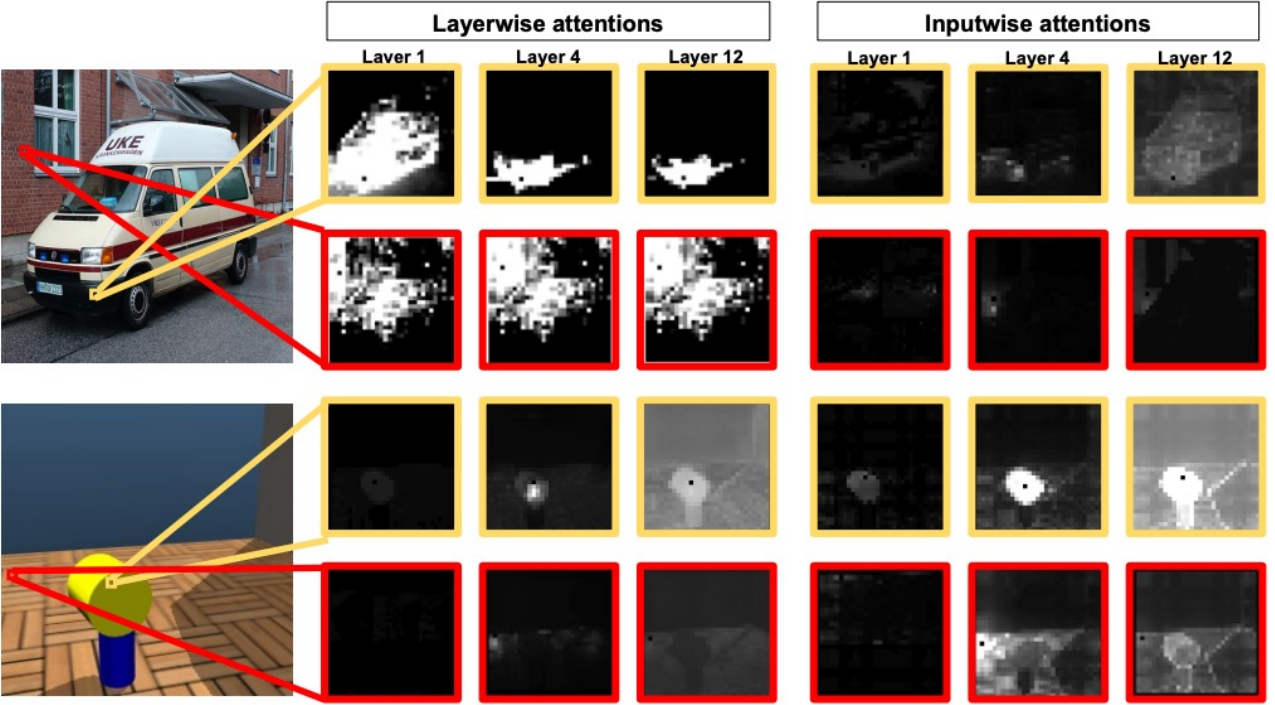}
\caption{Visualizing attentions for two images from ImageNet and Shapestacks. On each image, we consider a foreground patch (yellow) and a background patch (red). Attentions $a_{ii}$ for each patch towards itself are zeroed out to identify the patch in attention images. Layerwise attentions at various DINO layers (left) reveal little information at higher layers, so we instead compute inputwise attentions (right) using attention flow. }
\label{fig:layerwise-attentions}
\label{fig:visualizing-attentions}
\end{figure*}

At a high level, given an input image, our approach identifies \textit{attention-based} patch token groupings corresponding to object-like entities, and then computes \textit{activation-based} representations to describe each such identified object. We call our method Scene Objects From Transformers, or \soft for short. \soft acts as a wrapper around pre-trained vision transformer (\pvt) models. Rather than producing the standard activation-based ``scene vector'' representations from a \pvt, \softpvt produces an object-centric image representation that individuates, locates, and describes each object in the scene. This reflects the object-centric structure of the world~\cite{pmlr-v162-dittadi22a,yoon2023investigation,Greff2020-hq,van2019perspective}, which is useful for robotic manipulation~\cite{yuan2022sornet,jiang2022vima,zhu2023viola}.

Our experiments across simulated and real robotic manipulation settings demonstrate that: (1) \softpvt representations, which involve no new training, easily outperform vanilla \pvt activation features for 
manipulation for various choices of the backbone \pvt, (2) \softpvt with the best generic \pvt models largely bridge the gap to today's best robotics-specific image encoders,
and (3) \soft out-of-the-box consistently mines object groupings out of generic pre-trained vision transformer (\pvt) models in unseen robotics domains better than prior object-centric embedding approaches, even with domain-specific training.

\section{Related Work}
\paragraph{Robotics-Specific Pre-Trained Image Representations} In response to the growing evidence that widely used generic pre-trained image representations in computer vision form poor inputs for robotic policies, research has recently turned to robotics-specific image representations. Nair et al~\cite{nair2022r3m} first showed performance gains on various simulated and real robotics tasks from training time-contrastive representations on human videos~\cite{xu2022ego4d}. Since then, several other works~\cite{ma2022vip,ma2023liv,radosavovic2023real,majumdar2023we} have followed suit, demonstrating gains from various training objectives on robotics-relevant data~\cite{xu2022ego4d,damen2018scaling}, including objectives specifically motivated by usage for control tasks~\cite{ma2022vip,ma2023liv}. Recent works~\cite{hu2023pre,pumacay2024colosseum} come up with different benchmarks to study the effectiveness of pre-trained representations. However, the reason for the poor performance of generic models relative to these is not well-understood. We show that, at least for \textit{vision transformer} models, this gap may be largely bridged through changes to the representation inference procedure. 

\paragraph{Object-Centric Embeddings (OCE)} An OCE summarizes the scene in terms of the objects, their locations and properties, and their relationships to each other. A common OCE format for an image $x_i$ is $f_\theta(x_i)=(S=\{s_1, s_2, ..., s_{N_i}\}, C=\{c_1, c_2, ... c_{N_i}\})$.
This is a tuple of two unordered discrete sets, the object slot vectors $S$ and their coordinates $C$. Each slot vector $s_j\in \mathbb{R}^D$ describes one object or entity (such as an object part) in the scene, and the number $N_i$ of such slots varies from scene to scene. Each location specifier $c_i$ describes the location of its corresponding slot, such as through a segmentation mask, bounding box, or keypoint location. 
While the utility of acquiring such representations without manual supervision is widely recognized~\cite{pmlr-v162-dittadi22a,yoon2023investigation,Greff2020-hq,van2019perspective}, including for robotic manipulation~\cite{yuan2022sornet,jiang2022vima,zhu2023viola}, current approaches face several key challenges, most notably architecture design and optimization difficulties, which we elaborate upon in Appendix. At a high level, coaxing neural networks to generate unordered discrete sets of objects without explicit supervision requires architectural choices that create hyperparameter-sensitivity and other difficulties during optimization~\cite{Engelcke2020-rz,papa2022inductive,yang2022promising}. The end result is that even the best of these approaches struggle to generalize beyond narrow domains~\cite{singh2022neural,Sauvalle2022UnsupervisedMS,ISA,yang2022promising}. In contrast, our approach sidesteps all the above difficulties since it is only an inference procedure: it infers OCEs straight out of standard vision transformer architectures generating continuous scene-level activation maps and trained with mature and stable optimization procedures. As such, \softpvt can extract OCEs out-of-the-box on the same large natural image domains where the \pvt scene-level image encoders can operate.  

With the recent rise of pre-trained open-world object detectors~\cite{Kirillov2023SegmentA,Minderer2022SimpleOO}, several recent works have proposed to construct OCEs based on their output detections.  MOO~\cite{moo2023arxiv} uses OWL-ViT~\cite{Minderer2022SimpleOO} to infer centroid locations of foreground objects at the start of task execution. FOCUS~\cite{Ferraro2023FOCUSOW} uses SAM~\cite{Kirillov2023SegmentA} outputs to learn object-centric world models. GROOT~\cite{Zhu2023LearningGM} uses SAM segmentation masks to build 3D representations of task-relevant objects. Concurrent with our work, POCR~\cite{shi2024plug} builds OCEs by encoding SAM masks with various pre-trained vision encoders for robotics control.In contrast, our work is the first to construct useful OCEs from \textit{generic} pre-trained vision transformers such as DINO-v2~\cite{Oquab2023DINOv2LR} even when they are not explicitly trained for object recognition, by utilizing not only their outputs or final layer activations, but also their internal attentions.



\paragraph{Object Discovery from Image Encoders} Prior work has attempted to discover foreground objects from pre-trained convolutional or transformer image encoders by clustering pixel-wise or token-wise activations at one or more layers~\cite{Henaff2022-zv,LOST,tokencut,formula} using k-means or spectral clustering approaches. Recently, MaskCut~\cite{wang2023cut} apply such an approach iteratively to discover multiple objects, still using the pixel-wise activations. We compare against using such activation-clustering approaches in our experiments but find that they perform worse than our approach that uses a complementary source of information generated in a transformer's forward pass, namely, internal attentions, rather than key, value, query activations. 
DINO~\cite{caron2021emerging} first showed evidence that attention weights inferred within vision transformers may be informative for object discovery: its last layer CLS attentions cleanly segment the foreground objects in single-object images, a feature that is also employed in~\cite{Amir2021DeepVF}. MaskDistill~\cite{van2022discovering} builds on this, using feature similarities between the query and key features of the last DINO layer and the CLS token to identify a single foreground object, which is thereafter distilled into a Mask R-CNN encoder that can extract multiple objects. We propose an improved approach to use attentions for multi-object discovery in robotic manipulation scenes that requires no new training: it employs full token-token attention matrices rather than just CLS attentions, aggregates them across multiple layers using an ``attention rollout'' procedure, and exploits a handful of other images from the same domain, easily available in robotics settings, for background removal.

\begin{figure*}[t]
\centering
\includegraphics[width=0.8\textwidth]{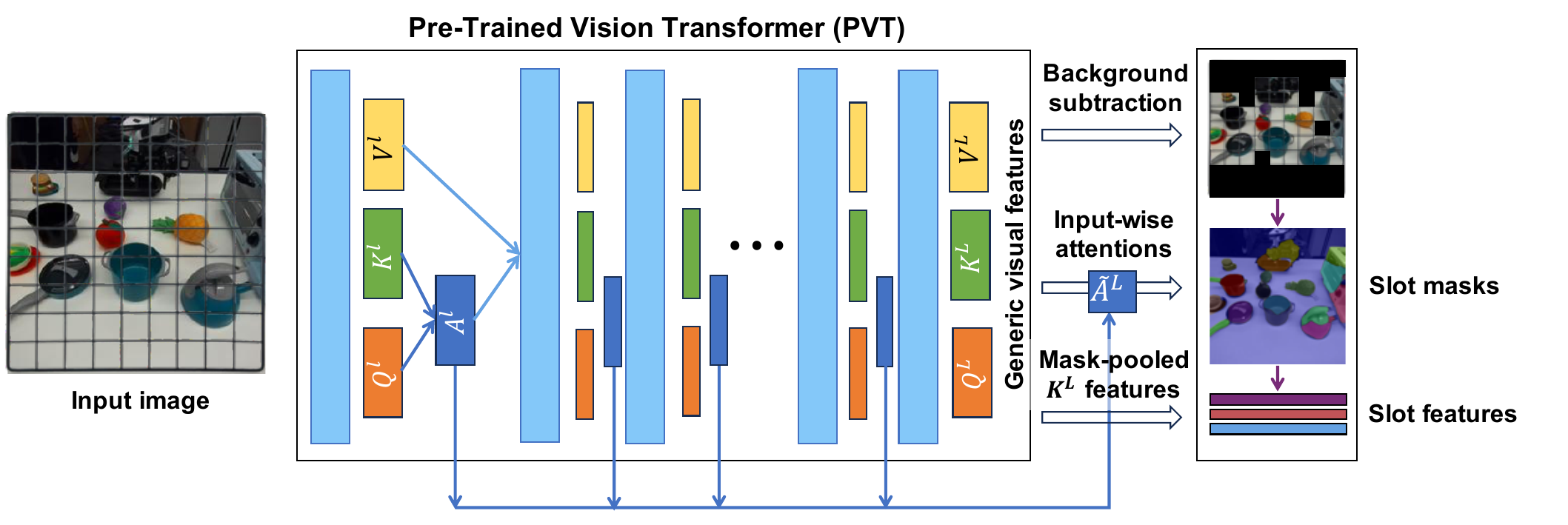}
\caption{\soft provides a wrapper around any pre-trained vision transformer model \pvt. Relying on nothing other than the activations and attentions throughout \pvt, \soft offers an alternative representation inference procedure to the standard last-layer activations. The resulting \softpvt representation is an object-centric image representation, suitable for off-the-shelf usage in robotic control tasks. $V^l$, $K^l$ and $Q^l$ represent the value, key and query of attention layers at layer $l$.}
\label{fig:soft-method-figure}
\label{fig:concept-fig}
\vspace{-0.5cm}
\end{figure*}

\section{Manipulation Policies Using Representations Inferred from Generic Vision Transformers}\label{sec:new-approach}


Consider the following setup. For learning a new manipulation task, we are provided with a small number $N$ of visual demonstration trajectories $\mathcal{D}=\{\tau_n\}_{n=1}^N$. The $n$-th trajectory is an image-action sequence $\tau_n=[x_{n,1}, u_{n,1}, x_{n,2}, u_{n,2}, ...]$. We would like to train a visual policy $\pi(x)$. For consistency with evaluations of prior pre-trained representations for robotics~\cite{nair2022r3m, ma2022vip, ma2023liv}, we will train policies with simple behavior cloning, minimizing the objective: $\pi^* = \min_\pi \sum_{n, t} \ell^2(\pi(x_{n,t}), u_{n,t})$ where $\ell$ is the squared $\ell^2$ error. 
We seek to develop a new pre-trained representation for robotics that would facilitate training such policies from limited demonstrations $\mathcal{D}$ to achieve higher rewards. Conceptually, we may compose a policy $\pi$ from image representation ($f_\psi)$ and action mapping ($g_\phi$) modules: $\pi(x)=g_\phi(f_\psi(x))$. With no pre-training, policy learning involves training both $f_\psi$ and $g_\phi$ from $\mathcal{D}$. With a pre-trained image encoder $f^*(\cdot)$, the policy $\pi(x)=g_\phi(f^*(x))$ has fewer learnable parameters, permitting learning from small $\mathcal{D}$. If the representations conveniently capture task-relevant information, the resulting policies would achieve high rewards. Good pre-trained representations must facilitate such sample-efficient, performant policy learning across a large range of unseen environments, objects, task, and robots.

We wish to develop a representation based on generic pre-trained vision transformers (\pvt)\cite{caron2021emerging, Oquab2023DINOv2LR, he2021masked, zhou2021ibot, touvron2020training}. 
As motivated above, standard feature activations at the output layers of such transformers have proven to be excellent pre-trained representations for vision tasks, but perform poorly for robotics; we verify this in Section~\ref{sec:exp}. How might we generate a better robotics-ready representation from a \pvt? To address this question, we will start by exploring \pvt self-attentions, whose utility as representations is ignored in prior work.


\subsection{The Information Held Within Transformer Self-Attentions}\label{sec:attention_viz} 
A vision transformer~\cite{dosovitskiy2020image} takes as input a serialized sequence of small $p$x$p$ image patches, e.g. $p=16$. An image with spatial dimensions $H$x$W$ is processed into $m=\frac{H}{p} \cdot \frac{W}{p}$ non-overlapping patch ``tokens''. Central to the computation in each layer $l$ of the transformer is the self-attention operator, which computes the next layer features for each token $y_i$ as a weighted average of the ``value'' features $v_j^l$ of other tokens $y_i^{l+1} \leftarrow \sum_j a_{ij}^l v_j^l$. These attention weights $a_{ij}^l$ are assigned by a function with learned parameters, and determine how much each other patch $j$, out of all $m$ patches, contributes to the new features of patch $i$. Information from all over the scene is thus mixed into the representations of each patch token. The ``attention matrix'' $A^l$ is an $m$x$m$ matrix, where the $i$-th row contains all $m$ attention weights for patch $i$. A more detailed description of the basic vision transformer architecture is provided in~\cite{dosovitskiy2020image}.

In the layers of a well-trained transformer, the attention weights $a_{ij}^l$ for a patch $i$, must be large for the patches $j$ that most provide context for understanding patch $i$. In particular, in images, the meanings of small $n$x$n$ tokens depend heavily on their context. For example, in the image in Fig~\ref{fig:layerwise-attentions} (top), no meaning can be assigned to the black patch token on the van bumper, without accounting for the context tokens, prominent among which is the rest of the van. The visualizations of attention weights from DINO~\cite{caron2021emerging} layers as $\frac{H}{n}$x$\frac{W}{n}$ grids, suggests that the weights might indeed capture information about object groupings. 

{How do these attention weights change from layer to layer, and does any one layer contain more information than others about the objects?} Fig~\ref{fig:layerwise-attentions} shows examples of attention weights over multiple layers. The attentions in the first layer are based on low-level color and position similarities between patches and do not yet capture object groupings, except in very simple scenes where objects have a single color or texture, easily distinguished from their backgrounds. For example, the bumper patch's attentions extend far beyond the van on the bottom left, yet they do not quite capture the top-right regions of the van. 
On the other hand, interpreting layerwise attentions becomes more complex in higher layers because these attentions operate over patch representations that are no longer based on purely local information; instead each patch representation at these layers already contains information from all over the scene.


Several methods have been proposed to resolve this mixing problem to produce more interpretable visualizations of attentions in later transformer layers~\cite{abnar2020quantifying, chefer2021transformer, ma2022visualizing}. We build on perhaps the simplest approach, ``attention rollout'', first presented by Abnar et al.~\cite{abnar2020quantifying} in the context of language transformers. We seek to know: how much do the features $y_i^l$ at the current layer $l$ of the transformer depend on the \textit{input} patch tokens $j$, which we will refer to as inputwise attentions $\tilde{a}_{ij}^l$. To answer this question, attention rollout prescribes matrix-multiplying the attention matrices $A^l$ of all preceding layers, corresponding to a linear model of information mixing across layers. To account for skip connections between layers, as in our models, we modify this~\cite{gil-vision-transformer-explainability} to:
\begin{equation}
\tilde{A}^l =  \prod_{l' \leq l} (A^{l'} + I).
\label{eq:rollout}
\end{equation}
To handle multi-head attentions, we simply average the attention matrices across the heads. Fig~\ref{fig:visualizing-attentions} visualizes inputwise attentions across various layers in DINO. 
Foreground patch attentions (yellow) in later layers are now more interpretable, and higher layers contain more clearly semantic object or entity grouping information. The formation of these input-wise attention matrices is schematically depicted in Figure~\ref{fig:concept-fig}.


\subsection{A Procedure To Infer Objects and Their Locations}

With these insights, we now propose a simple clustering-based procedure to permit \softpvt to individuate and locate objects. 
Given the inputwise attentions $\tilde{a}^l_{ij}$ of a vision transformer \pvt, \softpvt performs four steps:
\begin{itemize}[leftmargin=*]
\item \textbf{Symmetrized inputwise attentions as similarity measures}: \softpvt first computes inputwise attentions $\tilde{A}^L$ at the final layer $l=L$ as prescribed by Eq~\eqref{eq:rollout}. To explicitly account for object locations, we also compute a pairwise patch similarity matrix $S$. The symmetric matrix $A^s=\tilde{A}^L + \text{Transpose}(\tilde{A}^L) + S$ is treated as the measure of similarities between patch features. Both of the similarity matrices ($\tilde{A}^l + \text{Transpose}(\tilde{A}^L)$ and $S$) are normalized between 0 and 1.
In practice, we propagate only the top 10\% of attention weights in each layer to highlight the key regions.
\item  \textbf{Background removal}: While there is no domain-specific training set for producing \softpvt representations, the demonstration dataset $\mathcal{D}$, available in advance, does have other images within the domain. 
We sample $m=30$ reference frames per task from among the demonstrations $\mathcal{D}$ to identify and discard recurring, unchanging background regions. We follow the procedure described by Amir et al.~\cite{Amir2021DeepVF}: each patch token in the reference frames is assigned to foreground or background based on its CLS attention score. Then, each patch token in the current frame is assigned a backgroundness score based on matching its key features to these reference tokens. Scores higher than a threshold are discarded. More details in the appendix.
\item \textbf{Spectral clustering}: For the remaining tokens, we treat $A^*$ as the affinity matrix between tokens in spectral clustering~\cite{von2007tutorial}. \aqt{}The number of clusters may be automatically determined by the eigengap heuristic, permitting handling a varying number of objects in the scene.  
\item \textbf{Mask Refinement}: As described above, \soft assigns each $n$x$n$-pixel patch token to a cluster (8x8 in standard transformers), rather than each pixel. To generate smooth pixel-wise masks, we refine the patch-wise mask by applying a multi-label conditional random field~\cite{Krhenbhl2011EfficientII}. Details in appendix. 
\end{itemize}

\noindent Much of this procedure is visualized in Figure~\ref{fig:concept-fig}.

\subsection{Activation-Based Object Descriptions}

Having identified and located object-like regions as above, we now add slot vectors describing the contents of each region. For each object cluster ``slot'' $i$, 
its object slot vector $s_i$ is the average-pooled \pvt feature over the in-cluster patch tokens. At this stage, the combined object-centric scene descriptor \softpvt$(x)$ is an unordered set of slot vectors $\{s_i\}_{i=1}^{k}$, each corresponding to an object-like entity in the scene, and capturing its location and contents. The cardinality $k$, i.e., the number of object slots, varies by image. 


\subsection{Policy Learning from Demonstrations}

How might \softpvt be used as input to a policy, given that the slots are unordered and of variable number? While solutions based on set descriptors and graph neural networks are possible~\cite{kim2021transformers}, we opt for a simple solution based on explicit slot binding to achieve an ordered, fixed-size representation.
For each frame, we match \softpvt slots $\{s_{t,i}\}$ in the current frame to the averaged slot features from a randomly selected reference demonstration $\{\sum_j{s_{ref,j}}\}$. 
We use Hungarian matching~\cite{Kuhn1955TheHM} based on Euclidean distances between slot vectors, so that each reference slot is matched to at most
one slot in the current frame. Unmatched slots $s_{t,i}$ corresponding to spurious detections or distractor objects are discarded, and the remaining slots are now ordered as per the reference frame. 


Finally, we concatenate these newly ordered slot vectors $\{s_1, \dots, s_{k^*}\}_t$ into a flat, fixed-size representation ($k^*\times D$) and feed into a multi-layer perceptron $g_\phi$. 
\begin{equation}
    \pi(x_t) := g_\phi(\texttt{concat}(s_1(x_t), \dots, s_{k^*}(x_t)))
\end{equation}
This is now compatible with the behavior cloning policy training procedure described above.
A more formal pseudo-code description of \soft is provided in the supplementary material, and we will release all code upon publication. 

\section{Experiments}\label{sec:exp}
\soft provides an object-centric wrapper around pre-trained vision transformers (\pvt), aiming to improve the performance of generic pre-trained vision transformers as reusable representation encoders for robotic tasks. We first evaluate the extent to which \softpvt can successfully discover object-like regions in various robotic settings, out of the box. Next, we evaluate \softpvt object-centric embeddings to see: to what extent do they improve downstream policy performance compared to standard \pvt representations extracted from the same models? How does this compare to robotics-specific pre-trained representations?
\begin{figure}[t]
\centering
\includegraphics[width=0.9\linewidth]{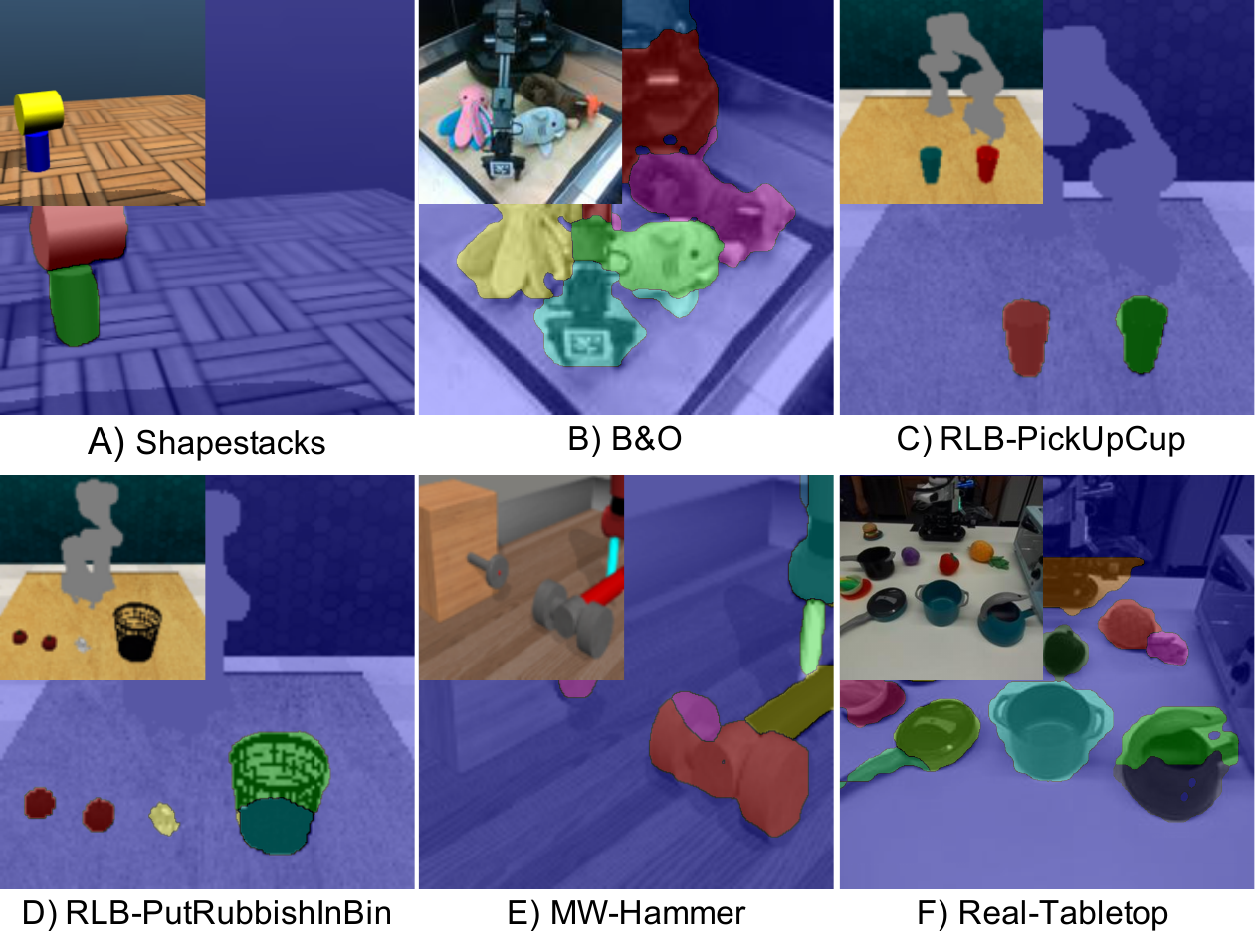}
\caption{Data from our datasets and simulation environments. We show the segmentation masks from \softpvt as well as the original images. Blue indicates background.}
\vspace{-0.5cm}
\label{fig:qual}
\end{figure}

\subsection{Discovering Object-Like Regions}

While \soft is not a segmentation method, it discovers object-like segmentation masks en route to generating a representation for robotics. We use these segments to evaluate \soft under various conditions. \\
\noindent \textbf{Datasets.} We first evaluate this component of \soft on 3 offline datasets with increasing degrees of difficulty: 
(1) \textbf{Shapestacks}~\cite{groth2018Shapestacks} is a synthetic dataset that has geometric shapes stacked on top of each other in various stable and unstable formations. 
(2) \textbf{RLB-PickUpCup} is a tabletop manipulation task in RLBench~\cite{james2020rlbench}, that we will use for control evaluations. \textbf{RLB-PickUpCup} requires the robot to pick up the red cup when presented with another distractor cup. We use the frames from 100 demos.
(3) \textbf{B\&O}~\cite{qian2022discovering} contains real images of a robot pushing deformable toy objects in its workspace, with annotated object masks.
All object locations are randomized at the start of each episode.\\
\noindent \textbf{\pvt backbones.} We evaluate \soft with various choices of \pvt backbones: 1) \textbf{DINO-ViT}~\cite{caron2021emerging}, a popular pretrained ViT network that produces useful representations for downstream vision and robotics tasks.  2) \textbf{DINOv2-ViT}~\cite{Oquab2023DINOv2LR} is the followup work that shows improved performances compared to DINO-VIT. 3) \textbf{DeiT}~\cite{deit,deit3} is a supervised ViT network trained on ImageNet-21k. 4) \textbf{ConvNext}~\cite{liu2022convnet} is a modern convolutional network that matches vision transformers on various vision benchmarks. Note that this is the only non-transformer based architecture, so we only apply the spectral clustering step and mask refinement step on the features of the last convolutional layer. 5) \textbf{MVP}~\cite{Xiao2022} is a transformer-based pretrained visual encoder for robot manipulation tasks.  Note that the background removal procedure only applies to ViT networks trained with unsupervised objectives, so we remove this step when performing experiments with DeiT and ConvNext and marked it as \texttt{{SOFT*}}.\\ 
\noindent \textbf{Activation-clustering.} With each backbone, we compare \soft's attention-based object discovery to \textbf{ODIN}~\cite{Henaff2022-zv}-style k-means clustering on final layer key features, and also iterative spectral clustering on key features as proposed in \textbf{MaskCut}~\cite{wang2023cut}. \\
\noindent \textbf{Unsupervised OCEs.} Finally, we compare against prior unsupervised object-centric embeddings, which typically train \textit{only} on in-domain images: \textbf{Genesis-V2}~\cite{Genesis-v2}, and the current state-of-the-art, \textbf{AST-SEG}~\cite{Sauvalle2022UnsupervisedMS}, unlike \soft which is \textit{only} pre-trained on out-of-domain images.
We report the standard segmentation metrics: Adjusted Rand Index (ARI)~\cite{ari1,ari2} and Mean Segmentation Covering (MSC)~\cite{arbelaez2010contour, engelcke2019genesis}, both in [0, 100], higher is better. See appendix for an explanation of these metrics. 
For each dataset, we compute the average metrics for each method over a test set of 320 images. 

\begin{table}[t]
     \caption{Segmentation metrics. Higher is better. 
     } 
  \label{tab:segmentation}
  \centering
  \resizebox{0.8\linewidth}{!}{
\begin{tabular}{c|c|cc|cc|cc}
\toprule
Backbone & Datasets$\rightarrow$ &\multicolumn{2}{c|}{Shapestacks} &\multicolumn{2}{c|}{RLB-PickUpCup}&\multicolumn{2}{c}{B\&O}\\
 & Num-images $\rightarrow$ & \multicolumn{2}{c}{20k} & \multicolumn{2}{c}{1.4k} & \multicolumn{2}{c}{300}\\
\midrule
 & Methods $\downarrow$  & ARI & MSC & ARI & MSC & ARI & MSC \\
\midrule
DINO-ViT &SOFT [Ours] & \textbf{0.80}& \textbf{0.86}  & \textbf{0.93} & \textbf{0.50} & \textbf{0.55} & \textbf{0.58} \\
&ODIN-feature & 0.39 &0.44 &0.20 &0.40 &0.54 &0.58 \\
&MaskCut\cite{wang2023cut} &0.08 & 0.24&0.447 &0.326  & 0.04&0.12 \\
\midrule
DINOv2-ViT &SOFT [Ours]& \textbf{0.72}& \textbf{0.65}& \textbf{0.99} &\textbf{0.51}  &\textbf{0.65} &\textbf{0.63} \\
&ODIN-feature& 0.04 &0.12  &0.00 & 0.22& 0.60&0.62 \\\midrule
DeiT &SOFT* [Ours] & 0.17 & 0.40  & \textbf{0.82} & \textbf{0.32}  & \textbf{0.69} & \textbf{0.52}\\
&ODIN-feature & 0.18 & 0.41  &0.08 &0.32  & 0.47 & 0.46\\
\midrule
ConvNext &SOFT* [Ours]  & 0.18 & 0.24  &0.23 &0.09& \textbf{0.42} & \textbf{0.30}\\
&ODIN-feature & 0.12 & 0.24  & 0.23&0.09   & 0.15 & 0.19 \\
\midrule
MVP~\cite{Xiao2022} & SOFT [Ours] & \textbf{0.69}& \textbf{0.55}&\textbf{0.80} &\textbf{0.43} &0.60 &0.53\\
&ODIN-feature &  0.67 & 0.44& 0.48&0.26 & 0.68&0.55   \\
\midrule
&Genesis-v2\cite{Genesis-v2} &0.73 &0.71 & -& -& 0.25 & 0.20  \\  
&AST-SEG\cite{Sauvalle2022UnsupervisedMS} & 0.73 &0.74 &- &- &0.03 &0.04 \\
\bottomrule
\end{tabular}
}
\end{table}
Table~\ref{tab:segmentation} shows these results, and Fig~\ref{fig:qual} shows examples of the segmentations masks on the evaluated datasets (more on website). 
First, with a fixed \pvt backbone, \softpvt object masks produced using our attention rollout procedure consistently match or outperform activation-clustering with ODIN or MaskCut, validating this choice.
Second, in-domain trained OCEs like Genesis-v2 or AST-SEG struggle to learn on smaller datasets, RLB-PickUpCup and B\&O. Finally, comparing among different backbones, DINO-ViT and DINOv2-ViT segments best match the object annotations.
\begin{table}[t]
\caption{Success Rates For Manipulation Tasks.}\label{tab:control}
\centering
  \resizebox{.85\linewidth}{!}{
  \begin{tabular}{c|c|c|c}
    \toprule
    & RLB-PickUpCup &RLB-PutRubbishInBin& MW-Hammer \\
    & 100 demos & 100 demos & 10 demos \\
    \midrule
    SOFT(DINOv2) [Ours] & 45 & 15 &73.4\\
    \midrule
    DINOv2-Flat-CLS & 10& 11&67.6\\
    DINOv2-Flat-Pool & 13.3& 5 & 69\\
    \midrule
    LIV~\cite{ma2023liv} & 49.3 & 21.9& 81.7\\
    R3M~\cite{nair2022r3m}  & 52.3 & 20.7& 81.3\\
    ImageNet & 14.0 & 6.7&53.7 \\
    \midrule
    SAM~\cite{sam} &15.9 & 4.4 & 51.5 \\
    \bottomrule
    \end{tabular}
}
\end{table}
\subsection{Robot Manipulation Experiments} 
We have thus far only evaluated \soft object masks, as a way to sanity-check our approach, and select the strongest \pvt backbone through cheap offline evaluations. We now evaluate \texttt{SOFT}(DINOv2) as an object-centric embedding for robot learning. We perform thorough experiments comparing various approaches quantitatively on two RLBench~\cite{james2020rlbench} and one Meta-World~\cite{Yu2019MetaWorldAB} tabletop manipulation environments: 
RLB-PickUpCup and RLB-PutRubbishInBin (see Figure~\ref{fig:qual}). In \textbf{RLB-PickUpCup}, the goal is to pick up the red cup in the presence of a distractor cup. The locations of the two cups are randomized for each episode as well as the color of the distractor. This requires the learned scene representations to include precise information about the cup's location and identity. During the task, cups could be heavily occluded by each other or the robot arm, thus the representation should also be robust under such conditions. In \textbf{RLB-PutRubbishInBin}, the robot needs to locate and pick up irregular-shaped rubbish among two other apples and put it in a trash can, all with randomized starting location. This is one of the more challenging tasks in RLBench, stress testing the representations. 
In \textbf{MW-Hammer}, the goal is to pick up a hammer and then smash a nail into a wall. The location of the hammer is randomized.

\noindent \textbf{Baselines.} Recall that \soft is a wrapper around a generic \pvt, in this case DINOv2. We compare this to using standard last layer activation-based features of the \pvt: (1) \textbf{DINOv2-Flat-CLS} uses the CLS token of the last attention layer, which is widely adopted as a representations for downstream tasks~\cite{caron2021emerging,Oquab2023DINOv2LR}, and 2) \textbf{DINOv2-Flat-Pool} mean-pools the key features over the entire image.  
Next, we compare to a CNN-based generic image feature, \textbf{ImageNet}-pretrained ResNet~\cite{he2016deep}, and to two state-of-the-art  
robotics-specific pre-trained features: LIV~\cite{ma2023liv} and R3M~\cite{nair2022r3m} that also use the ResNet backbone.
Finally, to evaluate whether good segmentation alone is sufficient, we compare against using the recent SOTA segmentation model, \textbf{SAM}~\cite{sam}. For this, we train a CNN encoder to process SAM object masks and feed into the policy. 

Table~\ref{tab:control} shows the success rates. The \texttt{SOFT}(DINOv2) OCE easily outperforms both the flat DINOv2 baselines, though they use the same backbone, trained with the same data. These more standard feature extraction approaches perform very poorly, as does ImageNet, which represents another standard representation for computer vision tasks.
It is precisely the poor performance of standard vision representations that prompted the recent development of more robotics-specific pre-trained representations such as LIV and R3M, which indeed perform much better. 
Here, \soft shines: using differently inferred representations from the same pre-trained vision transformer model DINOv2, it performs approximately on par with these robotics-aware representations.
Qualitatively, we observe that even when it fails, the learned \texttt{SOFT}(DINOv2) policy fails more gracefully than other methods, most often moving correctly to the right locations around the object of interest. See videos on website.

Finally, Figure~\ref{fig:robustness} studies the performance of \texttt{SOFT}(DINOv2) and DINOv2-Flat-CLS with varying numbers of training demonstrations: object-centric structure in \texttt{SOFT} enables learning from very few demonstrations.

\subsection{Real Robot Experiments}
We also evaluated \texttt{SOFT}(DINOv2) policies in a cluttered kitchen countertop setting (see Figure~\ref{fig:qual} F)), comparing them to DINOv2-Flat-CLS and LIV representations for the task of picking and placing toy fruits into a pot. We collect 50 expert demonstrations by teleoperation, and extract SOFT(DINOv2) features for policy learning. We randomized the placement of fruits during demo and evaluation. Our observations here are similar: DINOv2-CLS performs very poorly, but \texttt{SOFT}(DINOv2) performs noticeably better. With 10 trials of randomized placement of various fruits on the table, \texttt{SOFT}(DINOv2) has a success rate of 40\% which is as often as LIV policies. Figure~\ref{fig:success} shows a successful rollout of the policy learned with SOFT(DINOv2). We show videos of our sim and real robot trials on the website. 


\begin{figure}[t]
\centering
\includegraphics[width=1\linewidth]{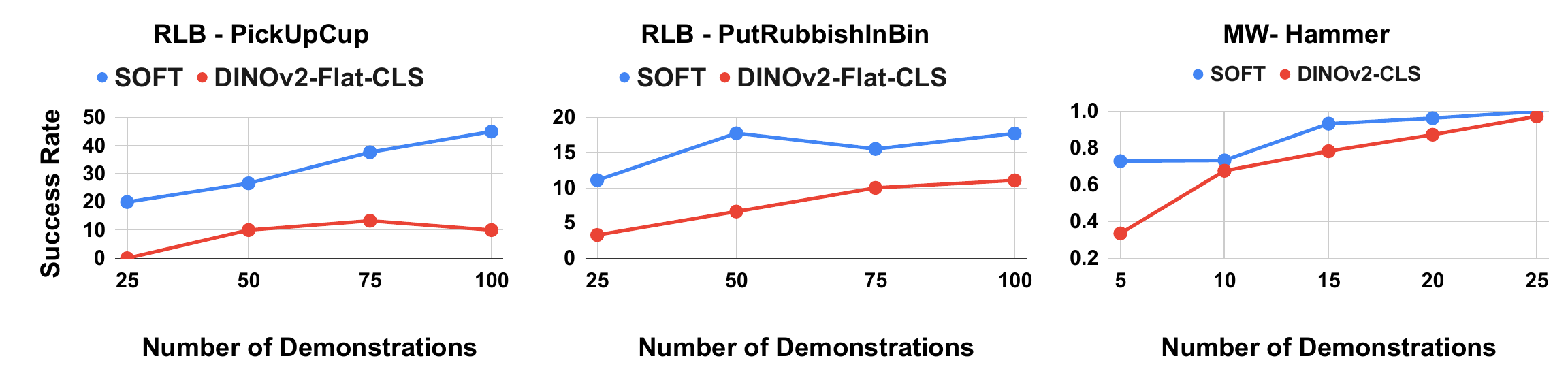}
\caption{Success rate of different methods as a function of the number of demonstrations.}
\label{fig:robustness}
\end{figure}

\begin{figure}[t]
\centering
\includegraphics[width=0.9\linewidth]{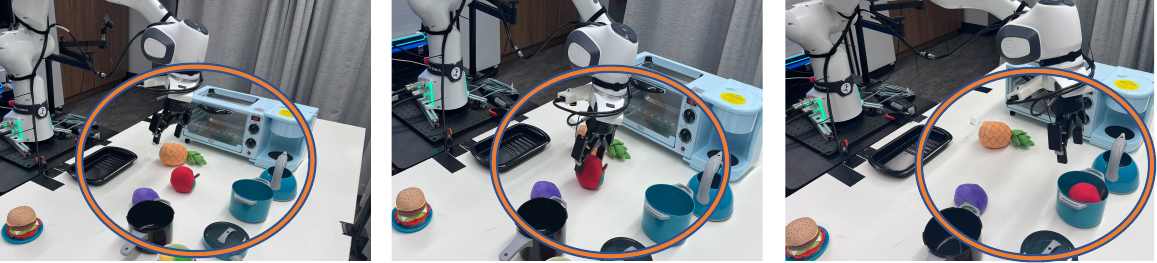}
\caption{Example of a successful policy rollout using SOFT(DINOv2)}
\vspace{-0.5cm}
\label{fig:success}
\end{figure}


\section{CONCLUSIONS}
By introducing a simple change to the representation inference procedure from a pre-trained vision transformer \pvt backbone, \softpvt produces object-centric embeddings that yield dramatic performance gains for control tasks, to the extent that it approximately matches the performance of the best robotics-specific pre-trained representations. Our results validate the utility of transformer attentions to inform representations and the power of object-centric embeddings, and demonstrate a route to making generic visual representations, wildly successful in computer vision, compatible with the specific needs of robot learners. 

\textbf{Acknowledgement:} This work is supported by NSF CAREER Award 2239301, ONR award N00014-22-1-2677 and by a gift from AWS AI for research in Trustworthy AI. 









(3) \textbf{B\&O}~\cite{qian2022discovering} contains real images of a robot pushing deformable toy objects in its workspace, with annotated object masks.
All object locations are randomized at the start of each episode.\\
\noindent \textbf{\pvt backbones.} We evaluate \soft with various choices of \pvt backbones: 1) \textbf{DINO-ViT}~\cite{caron2021emerging}, a popular pretrained ViT network that produces useful representations for downstream vision and robotics tasks.  2) \textbf{DINOv2-ViT}~\cite{Oquab2023DINOv2LR} is the followup work that shows improved performances compared to DINO-VIT. 3) \textbf{DeiT}~\cite{deit,deit3} is a supervised ViT network trained on ImageNet-21k. 4) \textbf{ConvNext}~\cite{liu2022convnet} is a modern convolutional network that matches vision transformers on various vision benchmarks. Note that this is the only non-transformer based architecture, so we only apply the spectral clustering step and mask refinement step on the features of the last convolutional layer. 5) \textbf{MVP}~\cite{Xiao2022} is a transformer-based pretrained visual encoder for robot manipulation tasks.  Note that the background removal procedure only applies to ViT networks trained with unsupervised objectives, so we remove this step when performing experiments with DeiT and ConvNext and marked it as \texttt{{SOFT*}}.\\ 
\noindent \textbf{Activation-clustering.} With each backbone, we compare \soft's attention-based object discovery to \textbf{ODIN}~\cite{Henaff2022-zv}-style k-means clustering on final layer key features, and also iterative spectral clustering on key features as proposed in \textbf{MaskCut}~\cite{wang2023cut}. \\
\noindent \textbf{Unsupervised OCEs.} Finally, we compare against prior unsupervised object-centric embeddings, which typically train \textit{only} on in-domain images: \textbf{Genesis-V2}~\cite{Genesis-v2}, and the current state-of-the-art, \textbf{AST-SEG}~\cite{Sauvalle2022UnsupervisedMS}, unlike \soft which is \textit{only} pre-trained on out-of-domain images.
We report the standard segmentation metrics: Adjusted Rand Index (ARI)~\cite{ari1,ari2} and Mean Segmentation Covering (MSC)~\cite{arbelaez2010contour, engelcke2019genesis}, both in [0, 100], higher is better. See appendix for an explanation of these metrics. 
For each dataset, we compute the average metrics for each method over a test set of 320 images. 

\begin{table}[t]
     \caption{Segmentation metrics. Higher is better. 
     } 
  \label{tab:segmentation}
  \centering
  \resizebox{0.8\linewidth}{!}{
\begin{tabular}{c|c|cc|cc|cc}
\toprule
Backbone & Datasets$\rightarrow$ &\multicolumn{2}{c|}{Shapestacks} &\multicolumn{2}{c|}{RLB-PickUpCup}&\multicolumn{2}{c}{B\&O}\\
 & Num-images $\rightarrow$ & \multicolumn{2}{c}{20k} & \multicolumn{2}{c}{1.4k} & \multicolumn{2}{c}{300}\\
\midrule
 & Methods $\downarrow$  & ARI & MSC & ARI & MSC & ARI & MSC \\
\midrule
DINO-ViT &SOFT [Ours] & \textbf{0.80}& \textbf{0.86}  & \textbf{0.93} & \textbf{0.50} & \textbf{0.55} & \textbf{0.58} \\
&ODIN-feature & 0.39 &0.44 &0.20 &0.40 &0.54 &0.58 \\
&MaskCut\cite{wang2023cut} &0.08 & 0.24&0.447 &0.326  & 0.04&0.12 \\
\midrule
DINOv2-ViT &SOFT [Ours]& \textbf{0.72}& \textbf{0.65}& \textbf{0.99} &\textbf{0.51}  &\textbf{0.65} &\textbf{0.63} \\
&ODIN-feature& 0.04 &0.12  &0.00 & 0.22& 0.60&0.62 \\\midrule
DeiT &SOFT* [Ours] & 0.17 & 0.40  & \textbf{0.82} & \textbf{0.32}  & \textbf{0.69} & \textbf{0.52}\\
&ODIN-feature & 0.18 & 0.41  &0.08 &0.32  & 0.47 & 0.46\\
\midrule
ConvNext &SOFT* [Ours]  & 0.18 & 0.24  &0.23 &0.09& \textbf{0.42} & \textbf{0.30}\\
&ODIN-feature & 0.12 & 0.24  & 0.23&0.09   & 0.15 & 0.19 \\
\midrule
MVP~\cite{Xiao2022} & SOFT [Ours] & \textbf{0.69}& \textbf{0.55}&\textbf{0.80} &\textbf{0.43} &0.60 &0.53\\
&ODIN-feature &  0.67 & 0.44& 0.48&0.26 & 0.68&0.55   \\
\midrule
&Genesis-v2\cite{Genesis-v2} &0.73 &0.71 & -& -& 0.25 & 0.20  \\  
&AST-SEG\cite{Sauvalle2022UnsupervisedMS} & 0.73 &0.74 &- &- &0.03 &0.04 \\
\bottomrule
\end{tabular}
}
\end{table}
Table~\ref{tab:segmentation} shows these results, and Fig~\ref{fig:qual} shows examples of the segmentations masks on the evaluated datasets (more on website). 
First, with a fixed \pvt backbone, \softpvt object masks produced using our attention rollout procedure consistently match or outperform activation-clustering with ODIN or MaskCut, validating this choice.
Second, in-domain trained OCEs like Genesis-v2 or AST-SEG struggle to learn on smaller datasets, RLB-PickUpCup and B\&O. Finally, comparing among different backbones, DINO-ViT and DINOv2-ViT segments best match the object annotations.
\begin{table}[t]
\caption{Success Rates For Manipulation Tasks.}\label{tab:control}
\centering
  \resizebox{.85\linewidth}{!}{
  \begin{tabular}{c|c|c|c}
    \toprule
    & RLB-PickUpCup &RLB-PutRubbishInBin& MW-Hammer \\
    & 100 demos & 100 demos & 10 demos \\
    \midrule
    SOFT(DINOv2) [Ours] & 45 & 15 &73.4\\
    \midrule
    DINOv2-Flat-CLS & 10& 11&67.6\\
    DINOv2-Flat-Pool & 13.3& 5 & 69\\
    \midrule
    LIV~\cite{ma2023liv} & 49.3 & 21.9& 81.7\\
    R3M~\cite{nair2022r3m}  & 52.3 & 20.7& 81.3\\
    ImageNet & 14.0 & 6.7&53.7 \\
    \midrule
    SAM~\cite{sam} &15.9 & 4.4 & 51.5 \\
    \bottomrule
    \end{tabular}
}
\end{table}
\subsection{Robot Manipulation Experiments} 
We have thus far only evaluated \soft object masks, as a way to sanity-check our approach, and select the strongest \pvt backbone through cheap offline evaluations. We now evaluate \texttt{SOFT}(DINOv2) as an object-centric embedding for robot learning. We perform thorough experiments comparing various approaches quantitatively on two RLBench~\cite{james2020rlbench} and one Meta-World~\cite{Yu2019MetaWorldAB} tabletop manipulation environments: 
RLB-PickUpCup and RLB-PutRubbishInBin (see Figure~\ref{fig:qual}). In \textbf{RLB-PickUpCup}, the goal is to pick up the red cup in the presence of a distractor cup. The locations of the two cups are randomized for each episode as well as the color of the distractor. This requires the learned scene representations to include precise information about the cup's location and identity. During the task, cups could be heavily occluded by each other or the robot arm, thus the representation should also be robust under such conditions. In \textbf{RLB-PutRubbishInBin}, the robot needs to locate and pick up irregular-shaped rubbish among two other apples and put it in a trash can, all with randomized starting location. This is one of the more challenging tasks in RLBench, stress testing the representations. 
In \textbf{MW-Hammer}, the goal is to pick up a hammer and then smash a nail into a wall. The location of the hammer is randomized.

\noindent \textbf{Baselines.} Recall that \soft is a wrapper around a generic \pvt, in this case DINOv2. We compare this to using standard last layer activation-based features of the \pvt: (1) \textbf{DINOv2-Flat-CLS} uses the CLS token of the last attention layer, which is widely adopted as a representations for downstream tasks~\cite{caron2021emerging,Oquab2023DINOv2LR}, and 2) \textbf{DINOv2-Flat-Pool} mean-pools the key features over the entire image.  
Next, we compare to a CNN-based generic image feature, \textbf{ImageNet}-pretrained ResNet~\cite{he2016deep}, and to two state-of-the-art  
robotics-specific pre-trained features: LIV~\cite{ma2023liv} and R3M~\cite{nair2022r3m} that also use the ResNet backbone.
Finally, to evaluate whether good segmentation alone is sufficient, we compare against using the recent SOTA segmentation model, \textbf{SAM}~\cite{sam}. For this, we train a CNN encoder to process SAM object masks and feed into the policy. 

Table~\ref{tab:control} shows the success rates. The \texttt{SOFT}(DINOv2) OCE easily outperforms both the flat DINOv2 baselines, though they use the same backbone, trained with the same data. These more standard feature extraction approaches perform very poorly, as does ImageNet, which represents another standard representation for computer vision tasks.
It is precisely the poor performance of standard vision representations that prompted the recent development of more robotics-specific pre-trained representations such as LIV and R3M, which indeed perform much better. 
Here, \soft shines: using differently inferred representations from the same pre-trained vision transformer model DINOv2, it performs approximately on par with these robotics-aware representations.
Qualitatively, we observe that even when it fails, the learned \texttt{SOFT}(DINOv2) policy fails more gracefully than other methods, most often moving correctly to the right locations around the object of interest. See videos on website.

Finally, Figure~\ref{fig:robustness} studies the performance of \texttt{SOFT}(DINOv2) and DINOv2-Flat-CLS with varying numbers of training demonstrations: object-centric structure in \texttt{SOFT} enables learning from very few demonstrations.

\subsection{Real Robot Experiments}
We also evaluated \texttt{SOFT}(DINOv2) policies in a cluttered kitchen countertop setting (see Figure~\ref{fig:qual} F)), comparing them to DINOv2-Flat-CLS and LIV representations for the task of picking and placing toy fruits into a pot. We collect 50 expert demonstrations by teleoperation, and extract SOFT(DINOv2) features for policy learning. We randomized the placement of fruits during demo and evaluation. Our observations here are similar: DINOv2-CLS performs very poorly, but \texttt{SOFT}(DINOv2) performs noticeably better. With 10 trials of randomized placement of various fruits on the table, \texttt{SOFT}(DINOv2) has a success rate of 40\% which is as often as LIV policies. Figure~\ref{fig:success} shows a successful rollout of the policy learned with SOFT(DINOv2). We show videos of our sim and real robot trials on the website. 


\begin{figure}[t]
\centering
\includegraphics[width=1\linewidth]{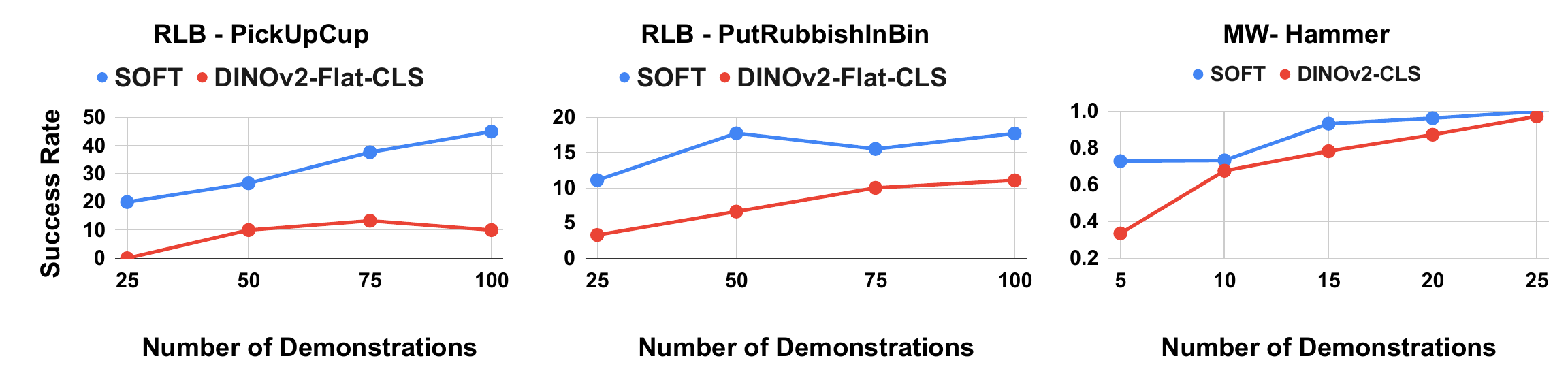}
\caption{Success rate of different methods as a function of the number of demonstrations.}
\label{fig:robustness}
\end{figure}

\begin{figure}[t]
\centering
\includegraphics[width=0.9\linewidth]{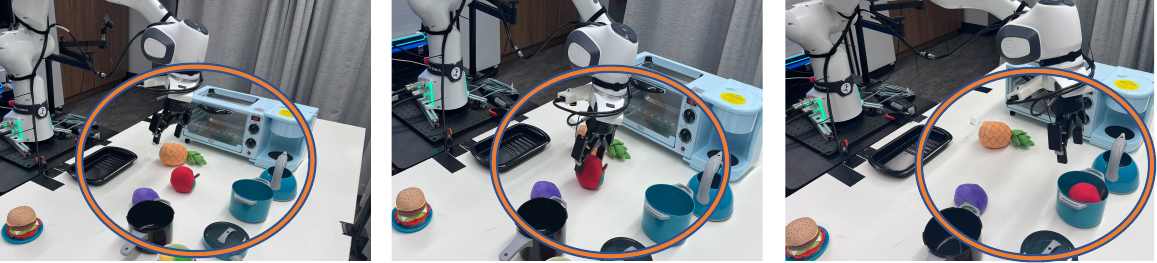}
\caption{Example of a successful policy rollout using SOFT(DINOv2)}
\vspace{-0.5cm}
\label{fig:success}
\end{figure}


\section{CONCLUSIONS}
By introducing a simple change to the representation inference procedure from a pre-trained vision transformer \pvt backbone, \softpvt produces object-centric embeddings that yield dramatic performance gains for control tasks, to the extent that it approximately matches the performance of the best robotics-specific pre-trained representations. Our results validate the utility of transformer attentions to inform representations and the power of object-centric embeddings, and demonstrate a route to making generic visual representations, wildly successful in computer vision, compatible with the specific needs of robot learners. 

\textbf{Acknowledgement:} This work is supported by NSF CAREER Award 2239301, ONR award N00014-22-1-2677 and by a gift from AWS AI for research in Trustworthy AI. 









\printbibliography
\end{document}


\maketitle
\thispagestyle{empty}
\pagestyle{empty}

\section{Key Challenges of Current Object-Centric Representation Approaches
}
Here we elaborate the challenges that most current OCR face:
\begin{enumerate}[leftmargin=*]
\item \textbf{Domain specificity:}
All OCR approaches even till date~\cite{singh2022neural,Sauvalle2022UnsupervisedMS,ISA} operate within highly restricted visual domains, usually containing varying configurations of a small number of synthetic object instances. For example, objects might be bright synthetic 2-D and 3-D geometric shapes with uniform colors or simple textures on a tabletop. For each such domain, OCR encoders must first be trained on training images / video from that domain before it can generate OCRs for held-out images. 
A few approaches (e.g. \cite{greff2019multi, jiang2019scalor, DeepLatentParticles, qian2022discovering}) have been evaluated on more naturalistic setups, such as images of faces or humans in varied poses, traffic scenes from a fixed road camera, overhead images of crowds or robot manipulation tabletop setups with simple objects. A recent study~\cite{yang2022promising} comprehensively documents the failings of unsupervised OCRs at segmenting real-world images. 
\item \textbf{Representation difficulties and architecture design:} OCRs contain \emph{unordered} sets of objects with \emph{varying cardinality}. These properties are not straightforward to generate in deep neural network representations. For handling \textbf{varying cardinality} $N_i$ of objects in images $x_i$, ``instance slot'' approaches~\cite{Kipf2019-hg, SA, DeepLatentParticles} conservatively generate $N_{max}$ object slot vectors at once to permit synthesizing any $N < N_{max}$ slots, but waste model capacity on unused slots and have to break the symmetry between slots. ``Spatial slot'' approaches~\cite{crawford2019spatially, jiang2019scalor, SPACE2020} go still further, associating each pixel with a slot vector and a position offset; this does not explicitly identify discrete objects in the scene and instead averages the ``objects'' prescribed by each pixel slot to generate the scene. ``Sequential slot'' approaches~\cite{greff2019multi, engelcke2019genesis} iteratively add objects until a termination condition, which causes slow inference and optimization difficulties.  
Next, most approaches ignore the \textbf{ordering invariance} within the OCR encoder, but \citet{Genesis-v2} use a stick-breaking process to achieve this, and some other works~\cite{SA, SAVi, SAVi++, Kipf2019-hg} build transformer or GNN-based \textit{decoder} models that treat input slot vectors as an unordered set so that the training objective, such as an autoencoding or prediction loss, effectively becomes invariant to slot order. 
\item \textbf{Resolution ambiguity:} The concept of ``objects'' involves natural ambiguity in resolution; the appropriate granularity of objects may vary from application to application. A pile of cloth might sometimes be best treated as one object, and sometimes as a set of individual garment objects. Most current OCR approaches have no control over the granularity of the learned objects. Recently, \citet{qian2022discovering} generate multiple levels of object granularity connected in a keypoint hierarchy, but there is once again no explicit control over the granularity of each level, and the hierarchy requires more complex architectures, objectives, and optimization.

\item \textbf{Optimization difficulties and inductive biases:} The relatively complex representation structure of OCRs also induces optimization difficulties. 
\citet{Engelcke2020-rz} observe that the expressive capacity of slot vectors is critical to the quality of the learned OCR: when it is chosen to be ``too small'' (such as through limited dimensionality), object information might not be effectively captured, and when it is too large, a single slot vector might already encode the entire scene, thus inducing no pressure to decompose the scene into multiple objects. 
More recently, \citet{papa2022inductive} expand these observations to note that OCRs are rather sensitive to many additional hyperparameters such as the architectural details in the encoder and the decoder and weighting terms in optimization objectives. These inductive biases are often heuristically picked, their effects on the learned representation are hard to predict, and when tuned for simple synthetic datasets as they often are, they transfer poorly to more realistic settings~\cite{yang2022promising}.

\end{enumerate}

\section{Segmentation Metrics}
On the benchmark datasets, we report quantitative results using standard metrics: adjusted rand index (ARI)~\cite{ari1,ari2} measures the fraction of times that predicted segments agree with ground truth segments on whether two pixels drawn from arbitrary ground truth segments belong on the same segment.
Mean segmentation covering (MSC)~\cite{arbelaez2010contour,engelcke2019genesis} explicitly matches each ground truth segment to a predicted segment and measures the mean of the pixel intersection over union for each match. Each method generates a single complete partitioning of the image patches into segments. 

\section{Mask Refinement Algorithm
}
For final mask refinement, we adopt the algorithm from~\cite{Amir2021DeepVF}, which uses multi-label CRF to refine the patch-wise masks. 
\newpage
\printbibliography